\documentclass{isprs} 
\usepackage{subcaption}
\usepackage{rotating}
\usepackage{setspace}
\usepackage{geometry} 
\usepackage{epstopdf}
\usepackage[labelsep=period]{caption} 
\usepackage[british]{babel} 
\usepackage[hang]{footmisc}
\usepackage{booktabs}
\usepackage{multirow}
 \usepackage{amsmath}
 \usepackage{amsfonts}
 \usepackage{placeins}

\begin{document}

\title{Spatially-Weighted CLIP for Street-View Geo-localization}

\author{
Ting Han\textsuperscript{1,2}, 
Fengjiao Li\textsuperscript{2}, 
Chunsong Chen\textsuperscript{1},
Haoling Huang\textsuperscript{3},
Yiping Chen\textsuperscript{1}$^{,*}$,
Meiliu Wu\textsuperscript{2}$^{,*}$}

\address{
	\textsuperscript{1}School of Geospatial Engineering and Science, Sun Yat-Sen University, Zhuhai, China\\
	\textsuperscript{2}School of Geographical and Earth Sciences, University of Glasgow, Glasgow, United Kingdom\\
    \textsuperscript{2}School of System Science and Engineering, Sun Yat-Sen University, Guangzhou, China\\
}

\maketitle

\sloppy

\section{Introduction}

Street-view geo-localization \cite{yan2024georeasoner,ye2025cross} aims to infer the geographic position of a query street-level image by matching it to a gallery of geo-tagged observations. In many real-world deployments, images are accompanied by GPS coordinates \cite{xu2024addressclip} but lack dense semantic text annotations, resulting in inherently weak supervision. Semantic text annotations (e.g., descriptions of streets, landmarks, or building functions) often provide richer localization cues than coordinates alone \cite{han2025towards}. By leveraging these two sources of information, we can view street-view geo-localization as learning retrieval-friendly representations under geographic constraints.

Specifically, we learn a shared embedding space in which a query image is closer to its true location and the descriptions within its spatial neighborhood, enabling similarity-based retrieval of the most plausible match from a reference database. Within this formulation, contrastive representation learning \cite{chen2020simple} is a natural choice, as it can directly exploit paired cross-modal relationships to learn retrieval-oriented embeddings. However, most contrastive objectives used in vision–language learning, especially CLIP-style InfoNCE \cite{radford2021learning,li2022blip}, implicitly assume a single discrete positive for each pair, treating all other candidates as equally negative. This assumption does not align with geographic space, where \textbf{\textit{nearby locations are inherently more related than distant ones}}.

Some CLIP-based efforts (e.g., GeoCLIP \cite{vivanco2023geoclip} and AddressCLIP \cite{xu2024addressclip}) in geospatial science show that CLIP-style retrieval can be adapted to geographic tags, such as GPS coordinates and address hierarchies. However, their contrastive supervision remains fundamentally pairwise: each query has a single positive, while all other samples in the mini-batch are treated as equally negative. As a result, the geographic “false-negative” issue is largely unresolved—locations that are spatially close and should be considered weak positives are instead pushed away as hard negatives during training. 

A defining property of geographic phenomena is spatial autocorrelation, commonly expressed by \textit{Tobler’s First Law of Geography (TFL): near things are more related than distant things} \cite{tobler1970computer}. For street-view images, the same principle suggests that observations captured within a neighborhood (e.g., the same block or adjacent intersections) should not be treated as uniformly negative relative to one another. Instead, they often exhibit continuity in both visual appearance and spatial context. If such nearby locations are penalized as hard negatives during training, the learned embedding space can become sensitive and spatially uncorrelated, leading to spatial bias, which means predictions that land in the correct vicinity but not exactly at the paired coordinate.

Spatial-aware CLIP \cite{han2025towards,suo2023text,wang2025spatialclip} variants highlight the importance of modeling spatial structure, but they do not explicitly instantiate Tobler-consistent, distance-decayed supervision for geo-localization. Therefore, our contribution is not “CLIP for geospatial tasks”, but a general mechanism for geographic alignment: distance-weighted soft supervision coupled with neighborhood-consistency regularization to preserve and strengthen neighborhood structure in the embedding space.

To this end, we propose \textbf{Spatially-Weighted CLIP (SW-CLIP)}, a novel spatial autocorrelation-oriented reformulation of CLIP that shifts the objective from semantic alignment to geographic alignment. The key idea is to (1) redefine the text branch as location-as-text, where a location caption encodes a real geographic position using coordinate tokens and administrative tokens, and (2) replace the one-hot supervision in InfoNCE with spatially weighted soft labels derived from geodesic distance. In addition, we explicitly enforce neighborhood consistency within each modality: nearby images should have similar image embeddings, and nearby location texts should have similar text embeddings. These components jointly inject spatial thinking into contrastive learning, producing cross-modal representations that preserve neighborhood structure and better support street-view geo-localization.

\section{Method}\label{sec:method}

\begin{figure*}
    \centering
    \includegraphics[width=0.95\linewidth]{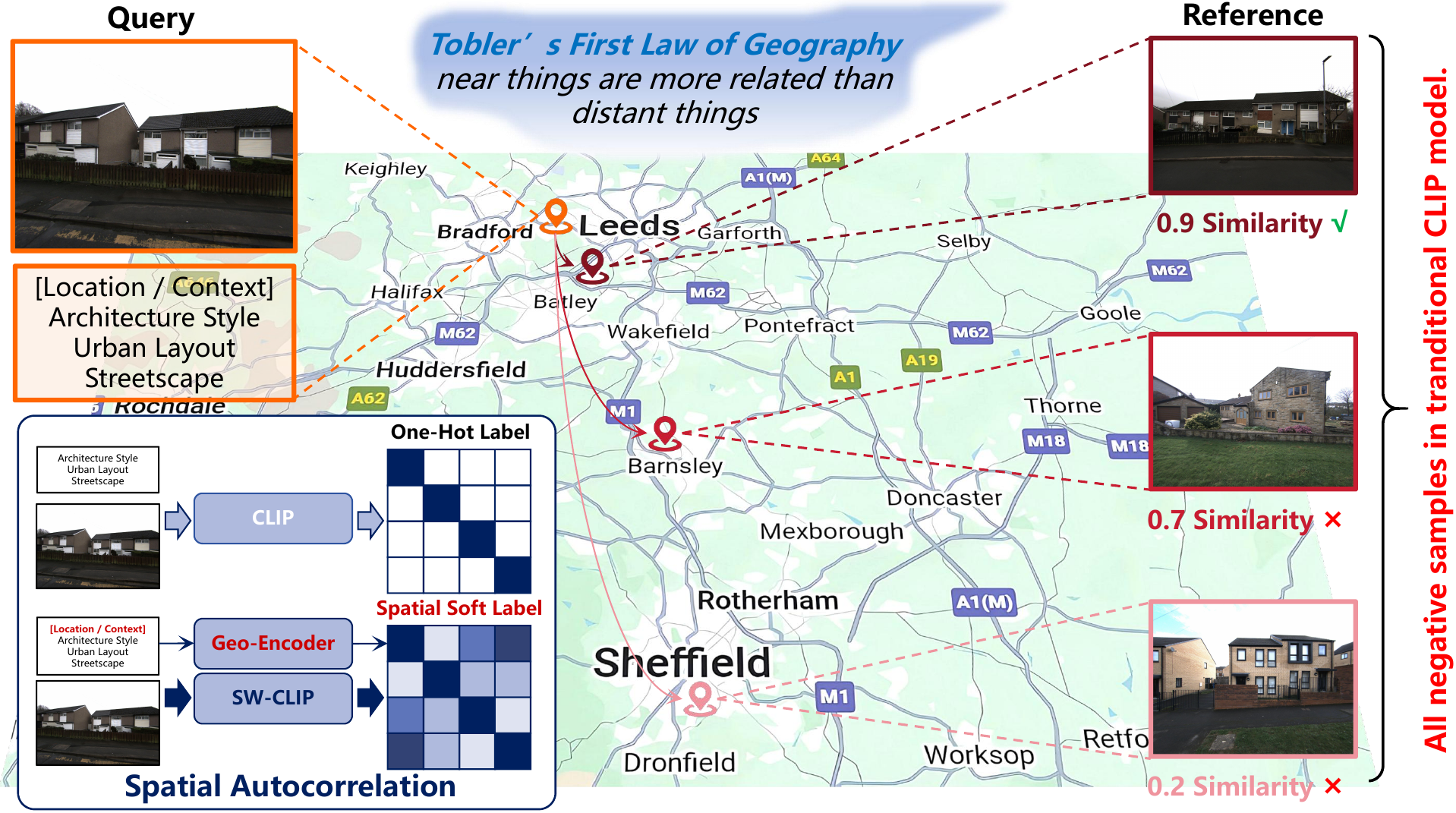}
    \caption{\textbf{Motivation and overview of SW-CLIP.} Standard CLIP training treats all non-matching samples in a mini-batch as equally negative, which can incorrectly penalize geographically nearby observations that share similar scene context. Guided by \textit{Tobler’s First Law of Geography and spatial autocorrelation}, SW-CLIP replaces the hard one-hot supervision with a distance-aware spatial soft label: nearby locations receive higher similarity targets while distant locations are down-weighted. This geographic weighting reduces false-negative conflicts in contrastive learning and encourages embeddings to be both retrieval-friendly and spatially coherent for street-view geo-localization.}
    \label{fig: motivation}
\end{figure*}

\subsection{Problem Definition}

We assume a dataset of geo-tagged street-view images ${(x_i, p_i)}^{N}_{i=1}$, where $x_i$ is an image and $p_i = (\phi_i, \lambda_i)$ denotes the latitude and longitude. Hierarchical identifiers such as street $s_i$ and city $c_i$ are also available. We learn an image encoder $F$ and a text encoder $G$ that map inputs into shared feature spaces.

We compute geodesic distances $d_{ij}$ using the numerically stable Haversine formula:
\begin{equation}
    d_{ij} = 2R arcsin(\sqrt{a_{ij}}),
\end{equation}
\begin{equation}
    a_{ij}= sin^{2}(\frac{\Delta\phi}{2})+cos\phi_{i}cos\phi_{j}sin^{2}(\frac{\Delta\lambda}{2}),
\end{equation}
where the Earth radius $R=6,371,000$m and angles are in radians. Distances provide the basis for spatial weighting and neighborhood definition.

\subsection{Location-as-Text}

Standard CLIP aligns images with natural-language captions that describe semantic content. For geo-localization, we instead define a location caption that explicitly encodes the real geographic position. A simple template is: [Street: $s_i$, City: $c_i$, Country: $k_i$. Lat: $\phi_i$, Lon: $\lambda_i$.]. The critical design choice is that the text branch becomes a learnable encoder of geographic position (Text features are not merely semantic; they are representations of a point in geographic space). Location-as-Text allows the model to retrieve locations using the same mechanism by which CLIP retrieves semantic captions, enabling within-text neighborhood constraints that encourage the location-text embedding space to a continuous geographic manifold.

Importantly, the key mechanism by which we inject spatial structure into learning is not the coordinate tokenization itself, but our distance-aware supervision. We compute pairwise geodesic distances $d_ij$ using the Haversine formula, derive a distance-decayed spatial soft-label distribution $w_ij$, and optimize a spatially-weighted InfoNCE objective so that nearby locations are treated as graded positives rather than uniformly hard negatives. Together with the neighborhood-consistency regularization described in our framework, these components encourage the embedding space to preserve Tobler-consistent neighborhood relationships.

\subsection{Spatially-Weighted CLIP}

\begin{table*}[t]
\centering
\small
\setlength{\tabcolsep}{6pt}
\renewcommand{\arraystretch}{1.15}
\caption{Main results of the proposed method in geo-localization task. The best results are in \textbf{bold}.}
\begin{tabular}{lcccccc}
\toprule
\textbf{Model} & \textbf{Med.\ GE (m)$\downarrow$} & \textbf{Mean\ GE (m)$\downarrow$} & \textbf{R@1$\uparrow$} & \textbf{Geo-Align$\uparrow$} & \textbf{SSI$\uparrow$} & \textbf{City-Align$\uparrow$} \\
\midrule
CLIP (ViT-B) & 276.80 & 6,723.54 & 0.421 & 0.000 & 0.103 & 0.259 \\
\midrule
SW-CLIP (ViT-B) & 91.96 & 449.25 & 0.910 & 0.550 & 0.621 & 0.737 \\
SW-CLIP (ViT-L) & \textbf{81.10} & \textbf{428.29} & \textbf{0.994} & \textbf{0.657} & \textbf{0.781} & \textbf{0.922} \\
\midrule
only L-a-T (ViT-B) & 185.40 & 2980.31 & 0.565 & 0.123 & 0.265 & 0.308 \\
\bottomrule
\end{tabular}
\label{tab:main_results}
\end{table*}

Given a batch of paired embeddings ${(v_i, t_i)}^{B}_{i=1}$, the standard CLIP image-to-text InfoNCE objective is:
\begin{equation}
    L_{CLIP} = -\frac{1}{B}\sum^{B}_{i=1}log\frac{exp(sim(v_i, t_i)/\tau)}{\sum^{B}_{j=1}exp(sim(v_i, t_j)/\tau)},
\end{equation}
where $\tau$ is a temperature. This defines exactly one positive per query and treats all other candidates as negatives.

Tobler's first law suggests that nearby locations are more closely related. We construct a spatial soft label distribution $w_{ij}$ over candidate location texts for each image query $i$. Firstly, we define a local distance-decay kernel:
\begin{equation}
    w^{local}_{ij} = exp(-\frac{d^{2}_{ij}}{2\sigma^{2}})\cdot \mathbf{1}[d_{ij} < d_{cut}],
\end{equation}
where $\sigma$ and $d_{cut}$ denote the scale and cutoff. Secondly, we incorporate hierarchical geographic relationships $w^{prior}_{ij}$ into normalized final weights $w_{ij}$. Finally, we modify InfoNCE by weighting the log-softmax term with the spatial soft labels:
\begin{equation}
    L_{SW} = -\frac{1}{B}\sum^{B}_{i=1}\sum^{B}_{j=1}w_{ij}log\frac{exp(sim(v_i, t_j)/\tau)}{\sum^{B}_{k=1}exp(sim(v_i, t_k)/\tau)},
\end{equation}
shifting CLIP from semantic alignment to geographic alignment. This strategy is rewarded for producing high similarity not only for the exact paired location text but also for geographically nearby location texts with smoothly decaying weights.

To account for spatial fairness, we introduce a region-level regularizer. We partition samples into $R$ geographic regions and denote by $B_r$ the set of indices in the current mini-batch that belong to region $r$. We define a differentiable regional performance proxy $\mathrm{Perf}_r$ and penalize its variance across regions:
\begin{equation}
L_{\mathrm{fair}} \;=\; \sum_{r=1}^{R}\Big(\mathrm{Perf}_r - \overline{\mathrm{Perf}}\Big)^2,
\overline{\mathrm{Perf}} \;=\; \frac{1}{R}\sum_{r=1}^{R}\mathrm{Perf}_r.
\label{eq:fair_var}
\end{equation}
where $\mathrm{Perf}_r$ denotes the average probability assigned to the ground-truth location text within the region $r$.

\section{Experiments}\label{sec:experiments}

\subsection{Configuration}

\subsubsection{Dataset} We use the xRI dataset\footnote{https://www.xri.online/} collected from eight cities in the United Kingdom, comprising 806 geo-referenced samples. Each sample consists of (1) an RGB street-view image and (2) the corresponding GPS coordinates in latitude and longitude. All samples are associated with a street block and city identifier using Claude-4.5.

\subsubsection{Metrics} We evaluate geo-localization as retrieval over a gallery of location texts with known coordinates. For each query image, we retrieve the top-K candidates by cosine similarity and compute the Geolocation error, Recall@K, Cross-modal alignment, and Spatial Smoothness Index. 

\subsubsection{Setups} We initialize from an off-the-shelf CLIP checkpoint with ViT-B/16 and ViT-L/14. All embeddings are L2-normalized, and similarity is computed via cosine similarity. We optimize the total loss $L_{total} = L_{SW} + \lambda_{fair}L_{fair}$, where $L_{fair}$ mitigates geographic imbalance across regions. We train using AdamW with weight decay and a cosine learning-rate schedule on a single RTX 3090 GPU. 

\subsection{Main Results}

Table~\ref{tab:main_results} summarizes the geo-localization performance of three settings: the CLIP baseline with a ViT-B backbone, our proposed SW-CLIP with the same backbone, and SW-CLIP with a larger ViT-H backbone. Injecting spatial autocorrelation into contrastive learning yields substantial gains in both localization accuracy and spatial coherence, while scaling up the backbone provides additional improvements.

Compared with CLIP (ViT-B), which achieves a median geo-localization error of 276.80 m and a mean error of 6,723.54 m, SW-CLIP (ViT-B) reduces the median error to 91.96 m and the mean error to 449.25 m. This indicates that spatially weighted supervision not only improves typical-case accuracy but also markedly suppresses long-tail failures that inflate the mean error. Retrieval performance improves dramatically under SW-CLIP, demonstrating that SW-CLIP primarily helps by consistently ranking the correct location, thereby reducing spatial bias.

Spatial structure diagnostics also confirm the effectiveness of our geographic alignment design. CLIP shows weak spatial consistency. In contrast, SW-CLIP substantially improves these metrics (Geo-Align, SSI, and City-Align), indicating more Tobler-consistent neighborhood smoothness and stronger city-level clustering in the embedding space. In summary, SW-CLIP consistently outperforms the CLIP baseline in geo-localization accuracy, retrieval reliability, and spatial coherence.

\subsection{Ablation Studies}

We report an ablation variant that uses Location-as-Text only (“only L-a-T”), i.e., replacing natural-language captions with location captions (street/city/country and coordinates) with standard CLIP InfoNCE loss. As shown in Table~\ref{tab:main_results}, this result yields a moderate improvement over CLIP, but remains substantially worse than SW-CLIP (ViT-B). Location prompting suggests limited neighborhood preservation in the embedding space. In contrast, SW-CLIP consistently boosts spatial coherence, validating our central claim: the key improvement comes from distance-weighted soft supervision and the resulting Tobler-consistent geographic alignment.

\section{Conclusion}\label{sec:conclusion}

We presented Spatially-Weighted CLIP (SW-CLIP), a geospatially inspired reformulation of CLIP for street-view geo-localization that explicitly injects spatial autocorrelation into contrastive learning. By encoding location as text and replacing one-hot InfoNCE supervision with distance-based soft labels, SW-CLIP learns embeddings that are geographically coherent. Experiments show large improvements in geolocation accuracy, with markedly higher spatial smoothness and city-level alignment. These results highlight the importance of moving from semantic alignment to geographic alignment for robust geo-localization and suggest a general recipe for incorporating spatial principles into multimodal representation learning.


{
	\begin{spacing}{1.17}
		\normalsize
		\bibliography{ISPRSguidelines_authors} 
	\end{spacing}
}

\end{document}